%% file: main.tex
\documentclass{article}
\usepackage{spconf,amsmath,epsfig}
\usepackage{booktabs}
\usepackage[table]{xcolor} 
\usepackage{physics}
\usepackage{amsmath}
\usepackage{tikz}
\usepackage{mathdots}
\usepackage{yhmath}
\usepackage{cancel}
\usepackage{color}
\usepackage{siunitx}
\usepackage{array}
\usepackage{multirow}
\usepackage{amssymb}
\usepackage{gensymb}
\usepackage{tabularx}
\usepackage{extarrows}
\usepackage{booktabs}
\usetikzlibrary{fadings}
\usetikzlibrary{patterns}
\usetikzlibrary{shadows.blur}
\usetikzlibrary{shapes}
\usepackage{tikz}
\usetikzlibrary{fadings}
\usetikzlibrary{patterns}
\usetikzlibrary{shadows.blur}
\usetikzlibrary{arrows,shapes,chains}

\usepackage{url}

\title{CLIP-Guided Source-Free Object Detection in Aerial Images}
\name{Nanqing Liu$^{1,2}$, Xun Xu$^2$, Yongyi Su$^2$, Chengxin Liu$^2$, Peiliang Gong$^2$, Heng-Chao Li$^1$}

\address{$^1$School of Information Science and Technology, Southwest Jiaotong University, PR China\\ $^2$Institute for Infocomm Research (I2R), A*STAR, Singapore}

\begin{document}

%
\maketitle
\maketitle
\begin{abstract}
Domain adaptation is crucial in aerial imagery, as the visual representation of these images can significantly vary based on factors such as geographic location, time, and weather conditions. Additionally, high-resolution aerial images often require substantial storage space and may not be readily accessible to the public. To address these challenges, we propose a novel Source-Free Object Detection (SFOD) method. Specifically, our approach begins with a self-training framework, which significantly enhances the performance of baseline methods. To alleviate the noisy labels in self-training, we utilize Contrastive Language-Image Pre-training (CLIP) to guide the generation of pseudo-labels, termed CLIP-guided Aggregation (CGA). By leveraging CLIP's zero-shot classification capability, we aggregate its scores with the original predicted bounding boxes, enabling us to obtain refined scores for the pseudo-labels. To validate the effectiveness of our method, we constructed two new datasets from different domains based on the DIOR dataset, named DIOR-C and DIOR-Cloudy. Experimental results demonstrate that our method outperforms other comparative algorithms. The code is available at \url{https://github.com/Lans1ng/SFOD-RS}.

\end{abstract}

\vspace{-0.3cm}
\section{Introduction}
\label{sec:intro}
In recent years, object detection in aerial imagery \cite{afdet,tinet,dior,oriented_rcnn,ossod} has seen growing interest due to its relevance in areas such as urban planning, environmental monitoring, and disaster management. Deep learning methods have been particularly successful in aerial object detection. However, these methods often exhibit limited generalization when applied to aerial images taken under various conditions, such as using different sensors or in different weather, leading to domain gaps or dataset biases.

To address these challenges, unsupervised domain adaptive object detection (UDAOD) has become a promising solution. Yet, UDAOD still depends on labeled data from the source domain, which poses a challenge in aerial imagery. High-resolution aerial images typically require substantial storage space and may not be easily accessible to the public. To overcome these issues, source-free object detection (SFOD)\cite{vs2023instance} has been introduced. This approach relies solely on a pre-trained source model and an unlabeled target dataset. Most current SFOD researches \cite{vs2023instance} are based on self-training methods\cite{ttac++,tribe}, like the mean-teacher framework \cite{ut}. In this setup, a teacher model guides a student model, but there is a risk of error accumulation, known as \textit{confirmation bias} if the teacher model provides incorrect learning targets.

Therefore, suppressing noisy labels generated during self-training is the key to solving this problem. Thanks to CLIP's \cite{clip} outstanding abilities in zero-shot learning and domain adaptation, we utilize it to assist in generating pseudo-label scores across different domains. It is worth noting that in few-shot object detection, VFA\cite{han2023vfa} also employs CLIP as an auxiliary for scoring predicted boxes. However, unlike it, we employ CLIP on the teacher's pseudo-labels to guide the student network's learning, which is termed CLIP-guided Aggregation (CGA). Specifically, the labels generated by CLIP are compared with the pseudo-labels from the teacher model. If they match, the original classification scores are retained; if they differ, the pseudo-label scores are adjusted through weighted means. This approach leverages the CLIP model as an anchor in the learning process, helping to correct errors and reduce \textit{confirmation bias}. To validate the effectiveness of our method, it needs to be tested on aerial images from different domains. The recent DOTA-C and DOTA-Cloudy datasets  \cite{he2023robustness} introduced various corruptions to assess the robustness of detectors. However, this testing process can only be finished on DOTA servers and is relatively cumbersome. Therefore, based on this, we developed new datasets for different domains, namely DIOR-C and DIOR-Cloudy, derived from the DIOR dataset\cite{dior}.

The main contributions of this paper are summarized as follows:

\begin{itemize}

\item To the best of our knowledge, we are the first to explore the SFOD method for oriented objects in aerial images.

\item We integrate the CLIP into the teacher-student self-training pipeline, which somewhat alleviates the accumulation of errors in pseudo-labels.

\item  We develop two new datasets, DIOR-C and DIOR-Cloudy, for domain adaptation. Our method demonstrates improved performance on these datasets compared to other methods.

\end{itemize}

\begin{figure}[t]

	\centering
\resizebox{1.0\linewidth}{!}{\input{pic/pipeline.tex}}
          \vspace{-0.5cm}
	\caption{\footnotesize{Overall architecture of the proposed SFOD method. (a) Self-training framework. (b) CLIP-guided Aggregation.}}
          \vspace{-0.5cm}
\label{fig:pipeline}
\end{figure}
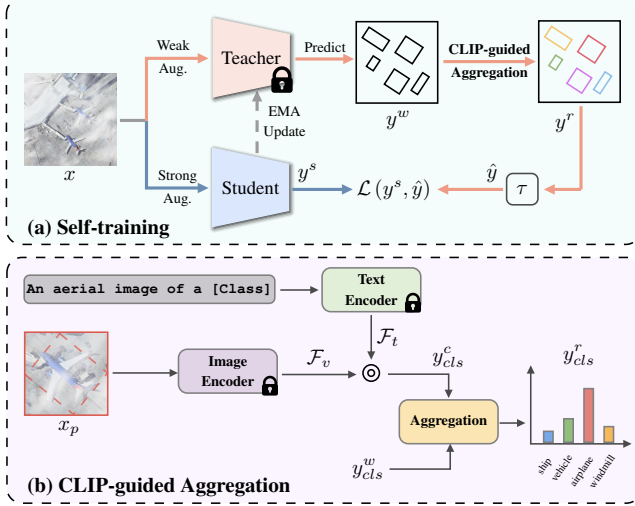

\vspace{-0.5cm}
\section{Methodology}
\label{sec:format}
In Source-Free Object Detection (SFOD), only the unlabeled target dataset $\mathcal{D}_t$ and a pre-trained source model $\Phi_p$ are utilized, without access to the labeled source dataset $\mathcal{D}_s$. As illustrated in Figure \ref{fig:pipeline}, we employ a self-training method specifically tailored for the target dataset \cite{ut}. In the pseudo-label generation phase, our method stabilizes pseudo-label scores by integrating CLIP-guided aggregation. This section will delve into these two key processes in detail.

\vspace{-0.2cm}
\subsection{Self-training}
As illustrated in Fig.\ref{fig:pipeline}(a), for a randomly selected target image $x$ from the dataset $\mathcal{D}_t$, we employ both weak and strong augmentation methods to generate $x^w$ and $x^s$, respectively. Weak augmentation is only horizontal flipping with a probability of 0.5. Strong augmentation includes a mix of color jittering, grayscale conversion, Gaussian blur, and Cutout. Both student and teacher models in our framework adopt the same network structure, specifically Oriented R-CNN \cite{oriented_rcnn}.

For the weakly augmented images $x^w$, we input them into the teacher model $\Phi_t$. Post-processing techniques such as Non-Maximum Suppression (NMS) are then applied to derive the classification scores $y_{cls}^w$ and regression parameters $y_{reg}^w$ for oriented objects. Recognizing the potential imprecision in initial object box scores, we implement a CLIP-guided Aggregation operation to refine the class scores, resulting in adjusted scores $y_{cls}^r$. The detailed methodology of this operation will be discussed in the following section. A confidence threshold $\tau$ is then used to filter out less probable predicted boxes, thus generating the final pseudo-labels $\hat{y} = \{\hat{y}_{cls}, \hat{y}_{reg}\}$.

In the parallel process for the strongly augmented images $x^s$, these are fed into the student model $\Phi_s$. This step similarly produces classification scores $y_{cls}^s$ and regression parameters $y_{reg}^s$ for oriented objects. The loss function is defined as:
\begin{equation}
\mathcal{L}=\mathcal{L}_{RoI}+\mathcal{L}_{RPN}
\end{equation}
where $\mathcal{L}_{RPN}$ and $\mathcal{L}_{RoI}$ represent the losses for the Region Proposal Network (RPN) and the Region of Interest (RoI) head, respectively.

To reduce the negative impact of inaccurate pseudo-labels, we update the teacher model through an exponential moving average (EMA) scheme, defined as $\Theta({\Phi_t}) \leftarrow \alpha \Theta({\Phi_t})+(1-\alpha) \Theta({\Phi_s})$, with the coefficient $\alpha$ set at 0.998.

\vspace{-0.3cm}
\subsection{CLIP-guided Aggregation}
To prevent the self-training task from being biased by incorrect pseudo labels, we utilize the zero-shot capabilities of CLIP to help assess the predicted results by the teacher model. Since the predicted aerial targets are in rotated boxes, to input these patches into CLIP, we first need to transform these targets into horizontal style. Assuming the parameters of a rotated box are ${x, y, w, h, \theta}$, we calculate the width ($w^{\prime}$) and height ($h^{\prime}$) of the corresponding horizontal bounding box using the following equations: $w^{\prime} = w \cdot|\cos \theta|+h \cdot|\sin \theta|, h^{\prime} = w \cdot|\sin \theta|+h \cdot|\cos \theta|$.

As shown in Fig.\ref{fig:pipeline}(b), we then use this horizontal bounding box to extract the relevant patch from the original image, denoted as $x_p \in \mathbb{R}^{N\times C\times H\times W}$. Here, $N$ represents the number of patches, while $C$, $H$, and $W$ denote the channels, height, and width, respectively. These patches $x_p$ are fed into CLIP’s image encoder to obtain the feature embeddings $\mathcal{F}_v \in \mathbb{R}^{N \times D}$, where $D$ is the dimensionality, typically 1024. Simultaneously, for the text branch, we formulate the text prompt simply as “$\texttt{An aerial image of a [Class]}$”, with $\texttt{[Class]}$ being a placeholder for the various classes in the dataset. These prompts are processed by CLIP’s pre-trained text encoder, resulting in embeddings $\mathcal{F}_t \in \mathbb{R}^{K \times D}$, where $K$ is the number of classes. The classification score $y^c_{cls} \in \mathbb{R}^{N \times K}$ for each patch is then calculated as $y^c_{cls} = \operatorname{Softmax}(\Vert \mathcal{F}_v \Vert \cdot \Vert \mathcal{F}_t^{\top} \Vert)$
\begin{table*}[ht]
    \scriptsize{}
    \caption{\footnotesize{Source-free domain adaptive object detection results on $\textbf{DIOR-C}$ and $\textbf{DIOR-Cloudy}$ dataset.}}
    \centering
    \setlength{\tabcolsep}{3pt} 
    \begin{tabular}{l|cccc|ccccc|ccccc|ccccc|c|c}
        \toprule
        \multirow{2}{*}{Model} & \multicolumn{19}{c|}{$\textbf{DIOR-C}$} & $\textbf{DIOR-Cloudy}$ & \multirow{2}{*}{mAP} \\
        \cmidrule{2-21}
        & Ga. & Shot & Im. & Spec. & De. & Glass & Mo. & Zoom & Ga. & Snow & Frost & Fog & Br. & Spat. & Co. & El. & Pixel & JPEG & Sa. & Cloudy & \\
        \midrule
        Clean & - & - & - & - & - & - & - & - & - & - & - & - & - & - & - & - & - & - & - & - & 54.1 \\
        \midrule
        Direct test & 13.1 & 12.2 & 13.6 & 15.6 & 28.5 & 24.4 & 29.4 & 18.6 & 29.6 & 22.8 & 25.8 & 36.3 & \textbf{49.9} & 34.0 & 32.7 & 43.7 & 40.1 & 47.5 & \textbf{52.6} & 39.0 & 30.5 \\
        Tent\cite{tent} & 15.8 & 14.6 & 16.4 & 16.6 & 19.5 & 16.4 & 21.9 & 11.9 & 20.6 & 16.3 & 18.2 & 26.8 & 28.5 & 21.6 & 27.5 & 26.1 & 26.5 & 25.8 & 30.4 & 23.6 & 21.3 \\
        BN\cite{bn} & 18.1 & 17.3 & 18.7 & 19.7 & 22.6 & 18.4 & 24.2 & 14.3 & 23.0 & 19.1 & 20.4 & 30.4 & 32.0 & 31.3 & 30.8 & 29.1 & 30.2 & 28.9 & 30.4 & 26.7 & 24.3 \\
        Shot\cite{shot} & 18.5 & 17.3 & 19.8 & 20.0 & 21.0 & 16.8 & 22.6 & 13.6 & 21.0 & 18.8 & 20.5 & 28.4 & 31.5 & 24.1 & 29.6 & 27.3 & 27.7 & 28.3 & 33.1 & 25.3 & 23.3 \\
        \rowcolor{orange!12}
        Self-training\cite{ut} & 24.1 & \textbf{21.1} & 24.2 & 25.0 & 36.4 & 35.4 & 39.7 & 27.8 & 39.1 & 30.0 & 31.9 & 47.7 & 47.9 & 38.8 & 47.5 & 45.6 & 45.9 & 46.7 & 50.1 & 42.1 & 37.4 \\
\rowcolor{blue!8}
CGA (Ours)&\textbf{25.6}&20.1&\textbf{26.0}&\textbf{25.4} &\textbf{37.9}&\textbf{35.9}&\textbf{40.2} &\textbf{30.3}&\textbf{40.1}&\textbf{31.5}&\textbf{32.7}&\textbf{50.0}&49.3&\textbf{39.8} &\textbf{47.8}& \textbf{47.3}&\textbf{47.0}& \textbf{47.6} &51.9 &\textbf{44.1}&38.5\\


\bottomrule
    \end{tabular}
    \label{tab:dior_exp}
    \vspace{-0.5cm}
\end{table*}

Suppose the predicted classification score of the teacher model is $y_{cls}^w \in \mathbb{R}^{N \times K}$. Then we can employ the CGA to derive refined scores $y_{cls}^r \in \mathbb{R}^{N \times K}$, which is defined as follows:
\begin{footnotesize}
\begin{equation}
y_{cls}^r= \begin{cases}y_{cls}^w, & \mkern-0mu\text { if } \operatorname{argmax}(y_{cls}^w) = \operatorname{argmax}(y_{cls}^c), \\ (1-\lambda) y_{cls}^w+\lambda y_{cls}^c, & \text { otherwise. }\end{cases}
\label{eq:lambda}
\end{equation}
\end{footnotesize}
In this equation, $\operatorname{argmax}$ represents the category index with the highest value. The coefficient $\lambda$ balances the original classification scores and those generated by CLIP. When the category predicted by CLIP matches the original score, indicating higher confidence in its accuracy, we keep the original score. Otherwise, we use a weighted combination of both scores as the final score. This method effectively filters out unstable category scores and works independently of the teacher-student learning cycle, reducing the chance of spreading incorrect labels.

\vspace{-0.5cm}
\section{Experiments}
\label{sec:pagestyle}
\subsection{Experimental Setup}
The datasets currently available for evaluating different aerial image domains include DOTA-C \cite{he2023robustness} and DOTA-Cloudy \cite{he2023robustness}. However, these datasets have certain limitations. Specifically, evaluations on them require the use of DOTA's server, and assessments for different types of corruption have to be conducted individually, posing significant challenges. To address this, we applied the corruptions from \cite{he2023robustness} to the DIOR dataset, resulting in two new datasets: DIOR-C and DIOR-Cloudy. DIOR-C incorporates 19 types of corruptions from ImageNet-C \cite{hendrycks2019robustness}, categorized into four groups: \textbf{Noise} (Gaussian, Shot, Impulse, Speckle), \textbf{Blur} (Defocus, Glass, Motion, Zoom, Gaussian), \textbf{Weather} (Snow, Frost, Fog, Brightness, Spatter), and \textbf{Digital} (Contrast, Elastic transform, Pixelate, JPEG compression, Saturate). For our experiments, we only generated images with a severity level of 3. DIOR-Cloudy is synthesized using publicly available cloudy images from the DOTA-Cloudy dataset. We used the original DIOR training set as our source data and introduced the above corruptions to the original DIOR validation and test sets. For specific experimental details, readers can refer to the code we have released.



\subsection{Source-Free Object Detection Results} 
We adopt the following generic state-of-the-art source-free domain adaptation methods to object detection tasks. \textbf{Direct test}: Directly apply the source model to test on the target dataset. \textbf{BN} \cite{bn}: Updates batch normalization statistics on the target data during testing. \textbf{Tent} \cite{tent}: Adapt a model by entropy minimization during testing. \textbf{Shot} \cite{shot}: Freeze the linear classiﬁcation head and train the target-speciﬁc feature extraction module. \textbf{Self training} \cite{ut}: Two kinds of augmentation methods applying to target data. The pseudo-label is generated by the teacher network and used to supervise the student network.

We assess the performance of these methods on $\textbf{DIOR-C}$ and $\textbf{DIOR-Cloudy}$, as detailed in Tab.\ref{tab:dior_exp}. The direct testing results show a significant decrease compared to the clean test (from 54.1$\%$ to 30.5$\%$), highlighting the substantial impact of image corruption. Notably, several source-free domain adaptive classification methods, such as Tent, BN, and Shot, demonstrate limited effectiveness when directly applied to detection tasks. The reason lies in classification methods often relying on statistics from large batch sizes for adaptation, which is not feasible in detection due to typically smaller batch sizes. Consequently, these methods sometimes perform even worse than direct testing. However, self-training significantly improved the results (from 30.5\% to 37.4\%), proving that it
is very effective for SFOD in aerial images. Our method CGA, by incorporating CLIP, can further enhance the effectiveness of self-training.
However, it is worth noting that our method does not show a significant improvement. This is because we only used the inference process of CLIP, and there is a gap between corrupted aerial images and CLIP. Therefore, its performance has not been fully exploited. In the future, we plan to decouple the text and image branches. This will allow us to adjust the scores for different texture categories accordingly.

\begin{figure}

    \centering
    \begin{tabular}{cc}
        \centering
        \includegraphics[width=0.4\linewidth]{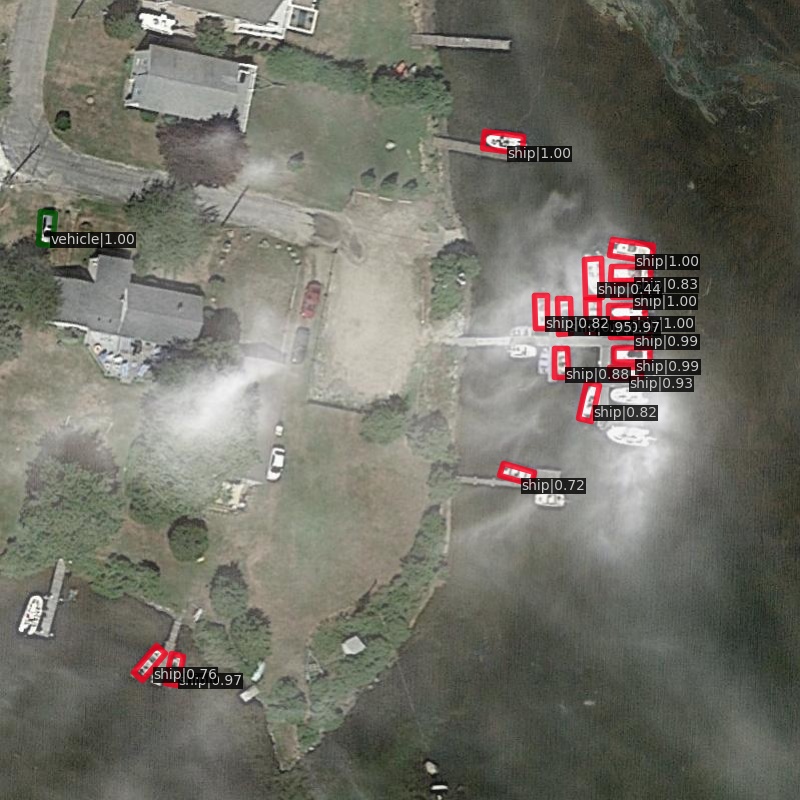} & \includegraphics[width=0.4\linewidth]{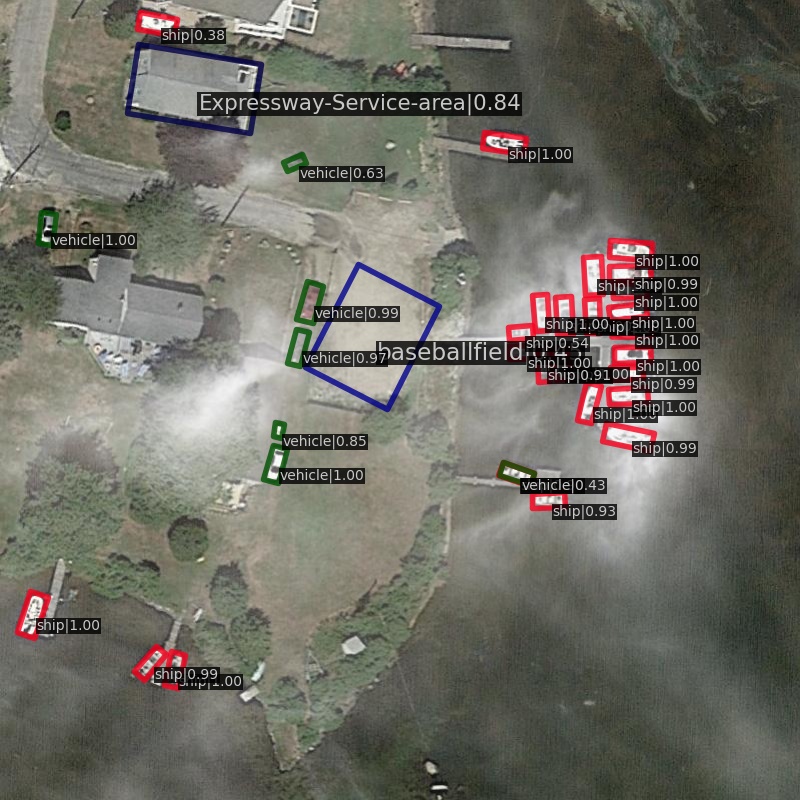} \\

        \footnotesize{(a) Direct test} & \footnotesize{(b) Self-training} \\
        \includegraphics[width=0.4\linewidth]{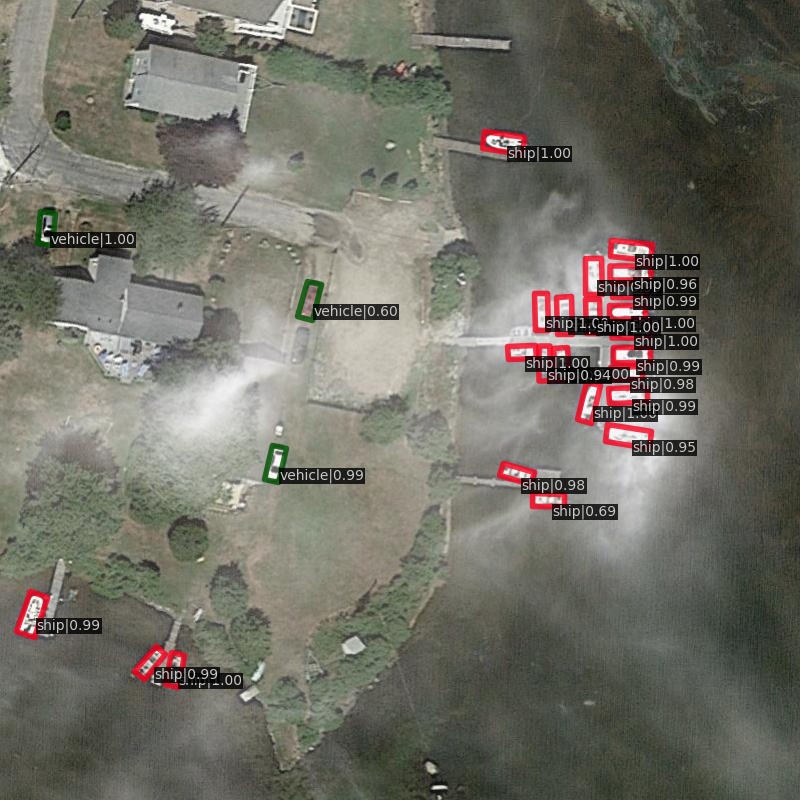} & \includegraphics[width=0.4\linewidth]{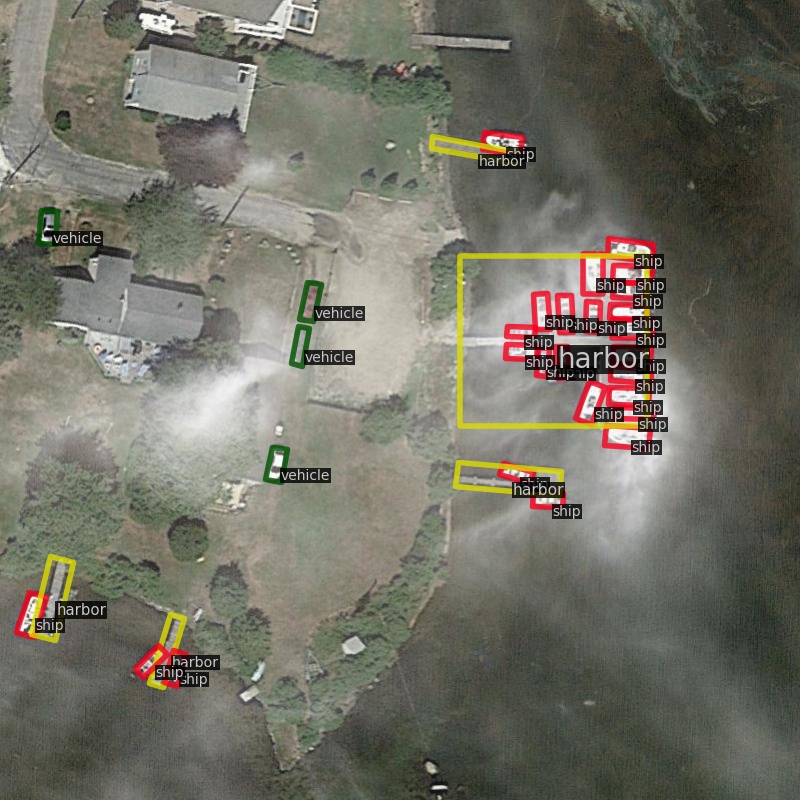} \\
        \footnotesize{(c) Ours} & \footnotesize{(d) Ground Truth} \\
    \end{tabular}
    \vspace{-0.3cm}
    \caption{\footnotesize{Qualitative results of different methods on DIOR-Cloudy dataset.}}
    \vspace{-0.5cm}
    \label{fig:vis}
\end{figure}

We also visualize the results of different methods in Fig. \ref{fig:vis}. It can be observed that the direct test method exhibits instances of missed detections such as "\textit{Vehicle}" and "\textit{Ship}" due to domain shift. Although the self-training method can detect the required samples, it also introduces many false alarms, such as "\textit{Baseball field}" and "\textit{Expressway-Service-area}" due to the accumulation of errors during the training process. All methods overlooked the "\textit{Harbor}" category in the detection results, but our method achieved the best performance.

\subsection{Ablation Study}
\begin{figure}[!htb]
	\flushleft
	
 	\begin{tabular}{cc}

	\hspace{-0.4cm}\includegraphics[width=0.5\linewidth]{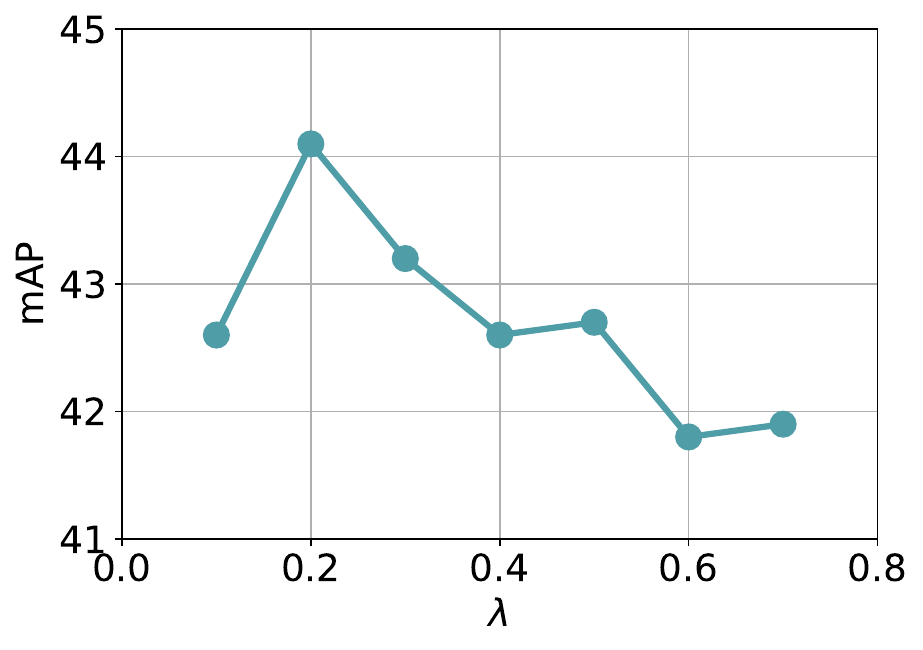}&\hspace{-0.5cm}	\includegraphics[width=0.5\linewidth]{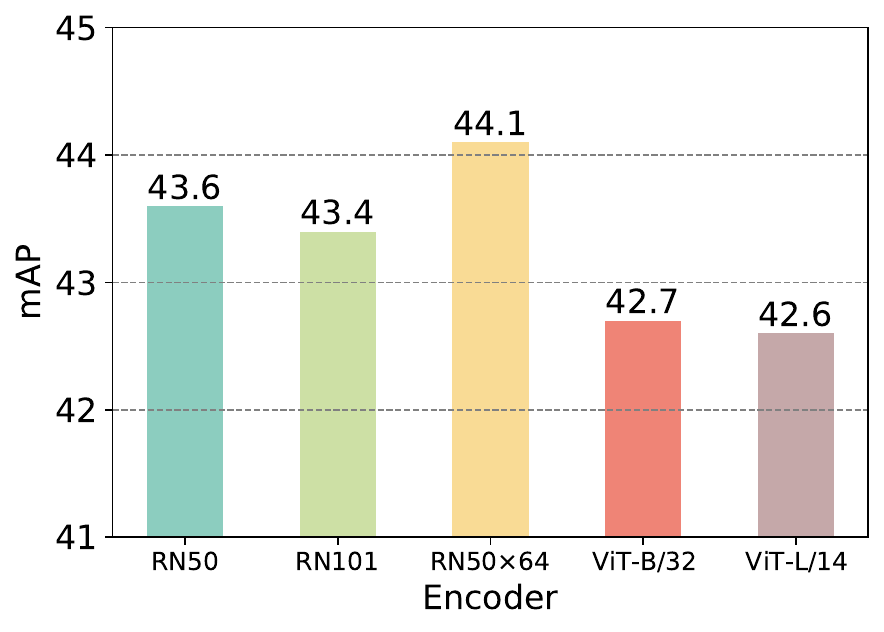}\\

	\end{tabular}
  \vspace{-0.4cm}
	\caption{\footnotesize{Ablation experiments about different $\lambda$ (left) and CLIP encoders (right) on the DIOR-Cloudy dataset.}}
	\label{fig:ablation}
 \vspace{-0.5cm}
\end{figure}

We firstly assess the impact of varying $\lambda$ values in Eq. \ref{eq:lambda}. As depicted in Fig.\ref{fig:ablation} (left), we observe that the optimal performance is achieved when $\lambda$ is set to 0.2. Increasing $\lambda$ further leads to a decline in effectiveness, indicating that excessive reliance on CLIP can negatively affect the network's ability to generate accurate pseudo-labels. Additionally, we examine the influence of different CLIP encoder structures, as shown in Fig.\ref{fig:ablation} (right). The results suggest that CNNs generally outperform transformers, mainly because the targets in aerial images are relatively simple. Consequently, transformers' global information processing capabilities do not provide substantial benefits in these scenarios.









\vspace{-0.3cm}
\section{Conclusion}
The paper introduces a new method for source-free object detection in aerial images, which integrates CLIP to guide the generation of pseudo-label scores. Experiments were conducted on specially created datasets, DIOR-C and DIOR-Cloudy, derived from the publicly available DIOR dataset. The proposed method is simple yet effective, outperforming comparative SFOD methods. However, we observed that CLIP's adaptability to certain types of corruption is limited. In the future, we plan to explore prompts designed specifically for different corruption to further improve performance.

\vspace{-0.2cm}
\bibliographystyle{IEEEbib}

\small
\bibliography{ref/refs}

\end{document}

%% file: pic/pipeline.tex
\tikzset{every picture/.style={line width=0.75pt}} 

\begin{tikzpicture}[x=0.75pt,y=0.75pt,yscale=-1,xscale=1]

\draw  [fill={rgb, 255:red, 144; green, 19; blue, 254 }  ,fill opacity=0.04 ][dash pattern={on 4.5pt off 4.5pt}] (60.67,185.93) .. controls (60.67,180.63) and (64.96,176.33) .. (70.27,176.33) -- (456.07,176.33) .. controls (461.37,176.33) and (465.67,180.63) .. (465.67,185.93) -- (465.67,323.73) .. controls (465.67,329.04) and (461.37,333.33) .. (456.07,333.33) -- (70.27,333.33) .. controls (64.96,333.33) and (60.67,329.04) .. (60.67,323.73) -- cycle ;
\draw  [fill={rgb, 255:red, 80; green, 227; blue, 194 }  ,fill opacity=0.06 ][dash pattern={on 4.5pt off 4.5pt}] (60.33,20.6) .. controls (60.33,15.3) and (64.63,11) .. (69.93,11) -- (455.73,11) .. controls (461.04,11) and (465.33,15.3) .. (465.33,20.6) -- (465.33,158.4) .. controls (465.33,163.7) and (461.04,168) .. (455.73,168) -- (69.93,168) .. controls (64.63,168) and (60.33,163.7) .. (60.33,158.4) -- cycle ;
\draw  [color={rgb, 255:red, 219; green, 231; blue, 253 }  ,draw opacity=1 ][fill={rgb, 255:red, 219; green, 231; blue, 253 }  ,fill opacity=1 ][blur shadow={shadow xshift=0.75pt,shadow yshift=-0.75pt, shadow blur radius=2.25pt, shadow blur steps=4 ,shadow opacity=100}] (192,106.57) -- (240.34,116.85) -- (240.34,146.29) -- (192,156.57) -- cycle ;
\draw [color={rgb, 255:red, 106; green, 140; blue, 178 }  ,draw opacity=1 ][line width=1.5]    (149.75,89) -- (149.75,131.96) -- (185.43,131.96) ;
\draw [shift={(189.43,131.96)}, rotate = 180] [fill={rgb, 255:red, 106; green, 140; blue, 178 }  ,fill opacity=1 ][line width=0.08]  [draw opacity=0] (6.97,-3.35) -- (0,0) -- (6.97,3.35) -- cycle    ;
\draw [color={rgb, 255:red, 155; green, 155; blue, 155 }  ,draw opacity=1 ][line width=1.5]  [dash pattern={on 5.63pt off 4.5pt}]  (220.36,109.69) -- (220.36,73.69) ;
\draw [shift={(220.36,69.69)}, rotate = 90] [fill={rgb, 255:red, 155; green, 155; blue, 155 }  ,fill opacity=1 ][line width=0.08]  [draw opacity=0] (6.97,-3.35) -- (0,0) -- (6.97,3.35) -- cycle    ;
\draw  [color={rgb, 255:red, 251; green, 229; blue, 224 }  ,draw opacity=1 ][fill={rgb, 255:red, 251; green, 229; blue, 224 }  ,fill opacity=1 ][blur shadow={shadow xshift=0.75pt,shadow yshift=-0.75pt, shadow blur radius=2.25pt, shadow blur steps=4 ,shadow opacity=100}] (192.47,23.69) -- (240.81,33.97) -- (240.81,63.42) -- (192.47,73.69) -- cycle ;
\draw (236.37,62.82) node  {\includegraphics[width=15pt,height=15pt]{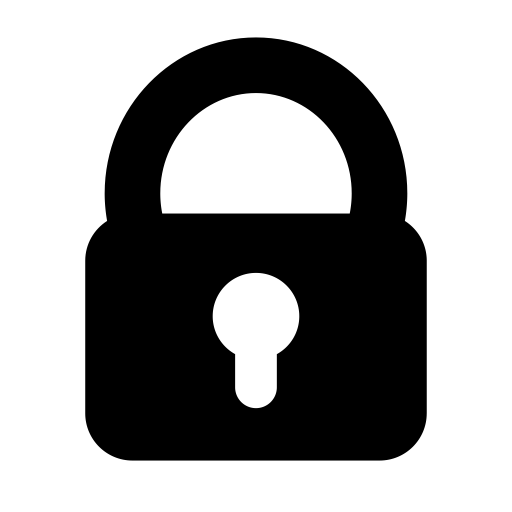}};
\draw [color={rgb, 255:red, 155; green, 155; blue, 155 }  ,draw opacity=1 ][line width=1.5]    (132.81,88.8) -- (149.75,88.8) ;
\draw [color={rgb, 255:red, 238; green, 156; blue, 133 }  ,draw opacity=1 ][line width=1.5]    (149.75,89) -- (149.75,49.11) -- (185.43,49.11) ;
\draw [shift={(189.43,49.11)}, rotate = 180] [fill={rgb, 255:red, 238; green, 156; blue, 133 }  ,fill opacity=1 ][line width=0.08]  [draw opacity=0] (6.97,-3.35) -- (0,0) -- (6.97,3.35) -- cycle    ;
\draw (101.26,87.93) node  {\includegraphics[width=44.14pt,height=44.14pt]{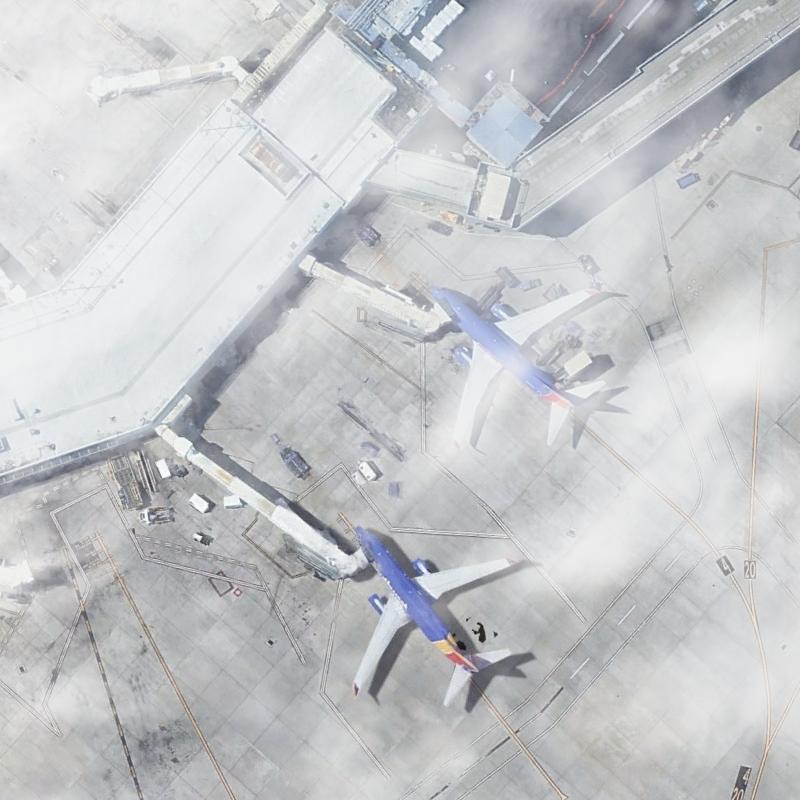}};
\draw  [color={rgb, 255:red, 128; green, 128; blue, 128 }  ,draw opacity=1 ][fill={rgb, 255:red, 155; green, 155; blue, 155 }  ,fill opacity=0.5 ] (71.67,191.22) .. controls (71.67,189.26) and (73.26,187.67) .. (75.22,187.67) -- (228.12,187.67) .. controls (230.08,187.67) and (231.67,189.26) .. (231.67,191.22) -- (231.67,202.12) .. controls (231.67,204.08) and (230.08,205.67) .. (228.12,205.67) -- (75.22,205.67) .. controls (73.26,205.67) and (71.67,204.08) .. (71.67,202.12) -- cycle ;
\draw  [color={rgb, 255:red, 128; green, 128; blue, 128 }  ,draw opacity=1 ][fill={rgb, 255:red, 226; green, 239; blue, 216 }  ,fill opacity=1 ] (261.86,188.59) .. controls (261.86,185.5) and (264.36,182.99) .. (267.46,182.99) -- (319.26,182.99) .. controls (322.35,182.99) and (324.86,185.5) .. (324.86,188.59) -- (324.86,205.39) .. controls (324.86,208.48) and (322.35,210.99) .. (319.26,210.99) -- (267.46,210.99) .. controls (264.36,210.99) and (261.86,208.48) .. (261.86,205.39) -- cycle ;
\draw (319.84,205.5) node  {\includegraphics[width=11.05pt,height=11.05pt]{lock.png}};
\draw  [color={rgb, 255:red, 128; green, 128; blue, 128 }  ,draw opacity=1 ][fill={rgb, 255:red, 225; green, 212; blue, 230 }  ,fill opacity=1 ] (170.19,240.59) .. controls (170.19,237.5) and (172.7,234.99) .. (175.79,234.99) -- (227.59,234.99) .. controls (230.68,234.99) and (233.19,237.5) .. (233.19,240.59) -- (233.19,257.39) .. controls (233.19,260.48) and (230.68,262.99) .. (227.59,262.99) -- (175.79,262.99) .. controls (172.7,262.99) and (170.19,260.48) .. (170.19,257.39) -- cycle ;
\draw (228.84,258.5) node  {\includegraphics[width=11.05pt,height=11.05pt]{lock.png}};
\draw [color={rgb, 255:red, 74; green, 74; blue, 74 }  ,draw opacity=1 ][line width=0.75]    (235.2,197.13) -- (256.67,197.13) ;
\draw [shift={(259.67,197.13)}, rotate = 180] [fill={rgb, 255:red, 74; green, 74; blue, 74 }  ,fill opacity=1 ][line width=0.08]  [draw opacity=0] (5.36,-2.57) -- (0,0) -- (5.36,2.57) -- cycle    ;
\draw [color={rgb, 255:red, 74; green, 74; blue, 74 }  ,draw opacity=1 ][line width=0.75]    (293.24,213.12) -- (293.24,238.69) ;
\draw [shift={(293.24,241.69)}, rotate = 270] [fill={rgb, 255:red, 74; green, 74; blue, 74 }  ,fill opacity=1 ][line width=0.08]  [draw opacity=0] (5.36,-2.57) -- (0,0) -- (5.36,2.57) -- cycle    ;
\draw   (291.11,250.17) .. controls (291.11,249.01) and (292.05,248.07) .. (293.21,248.07) .. controls (294.37,248.07) and (295.31,249.01) .. (295.31,250.17) .. controls (295.31,251.33) and (294.37,252.27) .. (293.21,252.27) .. controls (292.05,252.27) and (291.11,251.33) .. (291.11,250.17)(287.96,250.17) .. controls (287.96,247.27) and (290.31,244.92) .. (293.21,244.92) .. controls (296.11,244.92) and (298.46,247.27) .. (298.46,250.17) .. controls (298.46,253.07) and (296.11,255.42) .. (293.21,255.42) .. controls (290.31,255.42) and (287.96,253.07) .. (287.96,250.17) ;
\draw [color={rgb, 255:red, 74; green, 74; blue, 74 }  ,draw opacity=1 ][line width=0.75]    (303.93,250.92) -- (342.67,250.92) -- (342.67,263.19) ;
\draw [shift={(342.67,266.19)}, rotate = 270] [fill={rgb, 255:red, 74; green, 74; blue, 74 }  ,fill opacity=1 ][line width=0.08]  [draw opacity=0] (5.36,-2.57) -- (0,0) -- (5.36,2.57) -- cycle    ;
\draw   (283.67,25.69) -- (335.67,25.69) -- (335.67,77.69) -- (283.67,77.69) -- cycle ;
\draw   (308.01,55.91) -- (317.53,62.14) -- (311.15,71.86) -- (301.64,65.63) -- cycle ;
\draw   (315.33,34.97) -- (326.28,42.15) -- (319.44,52.58) -- (308.49,45.41) -- cycle ;
\draw   (326.08,58.27) -- (330.18,60.95) -- (322.79,72.22) -- (318.7,69.53) -- cycle ;
\draw   (291.87,29.23) -- (304.83,37.72) -- (300.45,44.39) -- (287.5,35.9) -- cycle ;
\draw   (294.53,47.56) -- (298.88,50.41) -- (294.51,57.08) -- (290.16,54.23) -- cycle ;

\draw [color={rgb, 255:red, 106; green, 140; blue, 178 }  ,draw opacity=1 ][line width=1.5]    (242.81,132.86) -- (277.48,132.86) ;
\draw [shift={(281.48,132.86)}, rotate = 180] [fill={rgb, 255:red, 106; green, 140; blue, 178 }  ,fill opacity=1 ][line width=0.08]  [draw opacity=0] (6.97,-3.35) -- (0,0) -- (6.97,3.35) -- cycle    ;
\draw [color={rgb, 255:red, 238; green, 156; blue, 133 }  ,draw opacity=1 ][line width=1.5]    (244.81,49.53) -- (278.33,49.53) ;
\draw [shift={(282.33,49.53)}, rotate = 180] [fill={rgb, 255:red, 238; green, 156; blue, 133 }  ,fill opacity=1 ][line width=0.08]  [draw opacity=0] (6.97,-3.35) -- (0,0) -- (6.97,3.35) -- cycle    ;
\draw [color={rgb, 255:red, 238; green, 156; blue, 133 }  ,draw opacity=1 ][line width=1.5]    (339.33,51.19) -- (394.33,51.19) ;
\draw [shift={(398.33,51.19)}, rotate = 180] [fill={rgb, 255:red, 238; green, 156; blue, 133 }  ,fill opacity=1 ][line width=0.08]  [draw opacity=0] (6.97,-3.35) -- (0,0) -- (6.97,3.35) -- cycle    ;
\draw [color={rgb, 255:red, 238; green, 156; blue, 133 }  ,draw opacity=1 ][line width=1.5]    (427.17,79.69) -- (427.17,89.69) -- (427.17,132.86) ;
\draw [color={rgb, 255:red, 238; green, 156; blue, 133 }  ,draw opacity=1 ][line width=1.5]    (377.43,132.86) -- (339.43,132.86) ;
\draw [shift={(335.43,132.86)}, rotate = 360] [fill={rgb, 255:red, 238; green, 156; blue, 133 }  ,fill opacity=1 ][line width=0.08]  [draw opacity=0] (6.97,-3.35) -- (0,0) -- (6.97,3.35) -- cycle    ;
\draw [color={rgb, 255:red, 74; green, 74; blue, 74 }  ,draw opacity=1 ][line width=0.75]    (304.93,311.56) -- (343.67,311.56) -- (343.67,298.19) ;
\draw [shift={(343.67,295.19)}, rotate = 90] [fill={rgb, 255:red, 74; green, 74; blue, 74 }  ,fill opacity=1 ][line width=0.08]  [draw opacity=0] (5.36,-2.57) -- (0,0) -- (5.36,2.57) -- cycle    ;
\draw  [color={rgb, 255:red, 128; green, 128; blue, 128 }  ,draw opacity=1 ][fill={rgb, 255:red, 254; green, 216; blue, 105 }  ,fill opacity=0.5 ] (311.19,272.92) .. controls (311.19,269.83) and (313.7,267.32) .. (316.79,267.32) -- (368.59,267.32) .. controls (371.68,267.32) and (374.19,269.83) .. (374.19,272.92) -- (374.19,289.72) .. controls (374.19,292.82) and (371.68,295.32) .. (368.59,295.32) -- (316.79,295.32) .. controls (313.7,295.32) and (311.19,292.82) .. (311.19,289.72) -- cycle ;
\draw [color={rgb, 255:red, 74; green, 74; blue, 74 }  ,draw opacity=1 ][line width=0.75]    (236.2,251.13) -- (280.87,251.13) ;
\draw [shift={(283.87,251.13)}, rotate = 180] [fill={rgb, 255:red, 74; green, 74; blue, 74 }  ,fill opacity=1 ][line width=0.08]  [draw opacity=0] (5.36,-2.57) -- (0,0) -- (5.36,2.57) -- cycle    ;
\draw  [color={rgb, 255:red, 74; green, 74; blue, 74 }  ,draw opacity=1 ] (379.67,126.69) .. controls (379.67,124.48) and (381.46,122.69) .. (383.67,122.69) -- (395.67,122.69) .. controls (397.88,122.69) and (399.67,124.48) .. (399.67,126.69) -- (399.67,138.69) .. controls (399.67,140.9) and (397.88,142.69) .. (395.67,142.69) -- (383.67,142.69) .. controls (381.46,142.69) and (379.67,140.9) .. (379.67,138.69) -- cycle ;
\draw [color={rgb, 255:red, 238; green, 156; blue, 133 }  ,draw opacity=1 ][line width=1.5]    (427.17,132.86) -- (407.43,132.86) ;
\draw [shift={(403.43,132.86)}, rotate = 360] [fill={rgb, 255:red, 238; green, 156; blue, 133 }  ,fill opacity=1 ][line width=0.08]  [draw opacity=0] (6.97,-3.35) -- (0,0) -- (6.97,3.35) -- cycle    ;
\draw [color={rgb, 255:red, 74; green, 74; blue, 74 }  ,draw opacity=1 ][line width=0.75]    (128.14,250.13) -- (166.87,250.13) ;
\draw [shift={(169.87,250.13)}, rotate = 180] [fill={rgb, 255:red, 74; green, 74; blue, 74 }  ,fill opacity=1 ][line width=0.08]  [draw opacity=0] (5.36,-2.57) -- (0,0) -- (5.36,2.57) -- cycle    ;
\draw [color={rgb, 255:red, 74; green, 74; blue, 74 }  ,draw opacity=1 ][line width=0.75]    (375.2,282.13) -- (388.67,282.13) ;
\draw [shift={(391.67,282.13)}, rotate = 180] [fill={rgb, 255:red, 74; green, 74; blue, 74 }  ,fill opacity=1 ][line width=0.08]  [draw opacity=0] (5.36,-2.57) -- (0,0) -- (5.36,2.57) -- cycle    ;
\draw [color={rgb, 255:red, 74; green, 74; blue, 74 }  ,draw opacity=1 ][line width=0.75]    (456.5,295.43) -- (395.08,295.43) -- (395.08,248.2) ;
\draw [shift={(395.08,245.2)}, rotate = 90] [fill={rgb, 255:red, 74; green, 74; blue, 74 }  ,fill opacity=1 ][line width=0.08]  [draw opacity=0] (5.36,-2.57) -- (0,0) -- (5.36,2.57) -- cycle    ;
\draw [shift={(459.5,295.43)}, rotate = 180] [fill={rgb, 255:red, 74; green, 74; blue, 74 }  ,fill opacity=1 ][line width=0.08]  [draw opacity=0] (5.36,-2.57) -- (0,0) -- (5.36,2.57) -- cycle    ;
\draw  [color={rgb, 255:red, 128; green, 128; blue, 128 }  ,draw opacity=1 ][fill={rgb, 255:red, 74; green, 144; blue, 226 }  ,fill opacity=0.76 ] (403.32,287.43) -- (409.32,287.43) -- (409.32,294.43) -- (403.32,294.43) -- cycle ;
\draw  [color={rgb, 255:red, 128; green, 128; blue, 128 }  ,draw opacity=1 ][fill={rgb, 255:red, 110; green, 172; blue, 78 }  ,fill opacity=0.75 ] (416.19,279.43) -- (422.19,279.43) -- (422.19,294.43) -- (416.19,294.43) -- cycle ;
\draw  [color={rgb, 255:red, 128; green, 128; blue, 128 }  ,draw opacity=1 ][fill={rgb, 255:red, 220; green, 89; blue, 79 }  ,fill opacity=0.75 ] (429.06,260.15) -- (435.06,260.15) -- (435.06,294.43) -- (429.06,294.43) -- cycle ;
\draw  [color={rgb, 255:red, 128; green, 128; blue, 128 }  ,draw opacity=1 ][fill={rgb, 255:red, 245; green, 166; blue, 35 }  ,fill opacity=0.75 ] (441.92,284.43) -- (447.92,284.43) -- (447.92,294.43) -- (441.92,294.43) -- cycle ;
\draw   (400.17,25.19) -- (452.17,25.19) -- (452.17,77.19) -- (400.17,77.19) -- cycle ;
\draw  [color={rgb, 255:red, 189; green, 16; blue, 224 }  ,draw opacity=0.56 ] (424.51,55.41) -- (434.03,61.64) -- (427.65,71.36) -- (418.14,65.13) -- cycle ;
\draw  [color={rgb, 255:red, 208; green, 2; blue, 27 }  ,draw opacity=0.61 ] (431.83,34.47) -- (442.78,41.65) -- (435.94,52.08) -- (424.99,44.91) -- cycle ;
\draw  [color={rgb, 255:red, 74; green, 144; blue, 226 }  ,draw opacity=0.59 ] (442.58,57.77) -- (446.68,60.45) -- (439.29,71.72) -- (435.2,69.03) -- cycle ;
\draw  [color={rgb, 255:red, 245; green, 166; blue, 35 }  ,draw opacity=0.64 ] (408.37,28.73) -- (421.33,37.22) -- (416.95,43.89) -- (404,35.4) -- cycle ;
\draw  [color={rgb, 255:red, 65; green, 117; blue, 5 }  ,draw opacity=0.52 ] (411.03,47.06) -- (415.38,49.91) -- (411.01,56.58) -- (406.66,53.73) -- cycle ;
\draw (98.5,250.6) node  {\includegraphics[width=39.75pt,height=39.75pt]{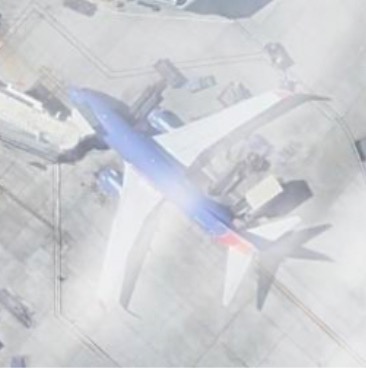}};
\draw  [color={rgb, 255:red, 220; green, 89; blue, 79 }  ,draw opacity=0.75 ][dash pattern={on 4.5pt off 4.5pt}][line width=0.75]  (94.41,225.01) -- (124.91,250.15) -- (103.12,276.59) -- (72.62,251.45) -- cycle ;
\draw  [color={rgb, 255:red, 220; green, 89; blue, 79 }  ,draw opacity=1 ][line width=0.75]  (72,224.1) -- (125,224.1) -- (125,277.1) -- (72,277.1) -- cycle ;

\draw (216.17,131.57) node   [align=left] {{\fontfamily{ptm}\selectfont {\small Student}}};
\draw (222.36,90.26) node [anchor=west] [inner sep=0.75pt]  [font=\footnotesize] [align=left] {\begin{minipage}[lt]{21.13pt}\setlength\topsep{0pt}
\begin{center}
{\scriptsize EMA}\\{\scriptsize Update}
\end{center}

\end{minipage}};
\draw (169.59,131.96) node  [font=\small] [align=left] {\begin{minipage}[lt]{23.44pt}\setlength\topsep{0pt}
\begin{center}
{\scriptsize Strong }\\{\scriptsize Aug.}
\end{center}

\end{minipage}};
\draw (166.8,49.11) node  [font=\small] [align=left] {\begin{minipage}[lt]{20.82pt}\setlength\topsep{0pt}
\begin{center}
{\scriptsize Weak }\\{\scriptsize Aug.}
\end{center}

\end{minipage}};
\draw (216.64,48.69) node   [align=left] {{\fontfamily{ptm}\selectfont {\small Teacher}}};
\draw (263.83,46.53) node [anchor=south] [inner sep=0.75pt]  [font=\small] [align=left] {\begin{minipage}[lt]{22.72pt}\setlength\topsep{0pt}
\begin{center}
{\scriptsize Predict}
\end{center}

\end{minipage}};
\draw (368.83,51.19) node  [font=\small] [align=left] {\begin{minipage}[lt]{42.35pt}\setlength\topsep{0pt}
\begin{center}
{\scriptsize \textbf{CLIP-guided }}\\{\scriptsize \textbf{Aggregation}}
\end{center}

\end{minipage}};
\draw (282.5,125.19) node [anchor=north west][inner sep=0.75pt]   [align=left] {$\displaystyle \mathcal{L}\left( y^{s} ,\hat{y}\right)$};
\draw (260.03,248.13) node [anchor=south] [inner sep=0.75pt]   [align=left] {$\displaystyle \mathcal{F}_{v}$};
\draw (295.24,227.41) node [anchor=west] [inner sep=0.75pt]   [align=left] {$\displaystyle \mathcal{F}_{t}$};
\draw (342.69,281.32) node  [font=\large] [align=left] {{\scriptsize \textbf{Aggregation}}};
\draw (201.69,248.99) node  [font=\large] [align=left] {\begin{minipage}[lt]{37pt}\setlength\topsep{0pt}
\begin{center}
{\scriptsize \textbf{Image }}\\\vspace{-0.15cm} {\scriptsize \textbf{Encoder}}
\end{center}

\end{minipage}};
\draw (389.67,132.69) node   [align=left] {$\displaystyle \tau $};
\draw (100.86,119.86) node [anchor=north] [inner sep=0.75pt]  [font=\normalsize] [align=left] {$\displaystyle x$};
\draw (375.43,129.86) node [anchor=south east] [inner sep=0.75pt]  [font=\normalsize] [align=left] {$\displaystyle \hat{y}$};
\draw (244.81,129.86) node [anchor=south west] [inner sep=0.75pt]  [font=\normalsize] [align=left] {$\displaystyle y^{s}$};
\draw (310.53,80.86) node [anchor=north] [inner sep=0.75pt]  [font=\normalsize] [align=left] {$\displaystyle y^{w}$};
\draw (417.37,320.75) node [anchor=north west][inner sep=0.75pt]  [font=\tiny,rotate=-303.99] [align=left] {airplane};
\draw (429.8,321.91) node [anchor=north west][inner sep=0.75pt]  [font=\tiny,rotate=-303.99] [align=left] {windmill};
\draw (405.43,317.88) node [anchor=north west][inner sep=0.75pt]  [font=\tiny,rotate=-303.99] [align=left] {vehicle};
\draw (396.45,313.08) node [anchor=north west][inner sep=0.75pt]  [font=\tiny,rotate=-303.99] [align=left] {ship};
\draw (342.67,247.92) node [anchor=south] [inner sep=0.75pt]   [align=left] {$\displaystyle y_{cls}^{c}$};
\draw (302.93,311.56) node [anchor=east] [inner sep=0.75pt]   [align=left] {$\displaystyle y_{cls}^{w}$};
\draw (426,246.59) node [anchor=south] [inner sep=0.75pt]   [align=left] {$\displaystyle y_{cls}^{r}$};
\draw (100.19,279.19) node [anchor=north] [inner sep=0.75pt]  [font=\normalsize] [align=left] {$\displaystyle x_{p}$};
\draw (425.17,82.69) node [anchor=north east] [inner sep=0.75pt]  [font=\normalsize] [align=left] {$\displaystyle y^{r}$};
\draw (72.27,165.5) node [anchor=south west] [inner sep=0.75pt]   [align=left] {\textbf{(a) Self-training}};
\draw (72.27,330.33) node [anchor=south west] [inner sep=0.75pt]   [align=left] {\textbf{(b) CLIP-guided Aggregation}};

\draw (293.36,196.99) node  [font=\large] [align=left] {\begin{minipage}[lt]{37pt}\setlength\topsep{0pt}
\begin{center}
{\scriptsize \textbf{Text }}\\\vspace{-0.15cm} {\scriptsize \textbf{Encoder}}
\end{center}

\end{minipage}};
\draw (151.67,196.67) node  [font=\scriptsize] [align=left] {{\fontfamily{pcr}\selectfont \textbf{An aerial image of a [Class]}}};

\end{tikzpicture}